\documentclass{article}
\usepackage{ijcai18}

\usepackage{times}
\usepackage{xcolor}
\usepackage{amsfonts}
\usepackage{soul}
\usepackage{algorithm}
\usepackage{multirow}
\usepackage{graphicx}
\usepackage{subfigure}
\usepackage{epstopdf}
\usepackage{multicol}
\usepackage{algorithmic}
\usepackage[utf8]{inputenc}
\usepackage[small]{caption}
\usepackage{array}

\title{EA-LSTM: Evolutionary Attention-based LSTM\\ for Time Series Prediction}

\author{
	Youru Li$^{\dag,\ddag}$, 
	Zhenfeng Zhu$^{\dag,\ddag}$, 
	Deqiang Kong$^\equiv$, 
	Hua Han$^{\Xi}$,
	Yao Zhao$^{\dag,\ddag}$,
	\\ 
	$^\dag$ Institute of Information Science, Beijing Jiaotong University, Beijing, China \\
	$^\ddag$ Beijing Key Laboratory of Advanced Information Science \\and Network Technology, Beijing, China\\
	$^\equiv$ Microsoft Multimedia, Beijing, China  \\
	$\Xi$ National Laboratory of Pattern Recognition, Institute of Automation, \\Chinese Academy of Sciences (CAS), Beijing, China\\
	$^{\dag,\ddag}$ \{liyouru,zhfzhu,xumx0721,yzhao\}@bjtu.edu.cn,\\
	$^\equiv$ kodeqian@microsoft.com
	$^\Xi$ hua.han@ia.ac.cn
}
\begin{document}

\maketitle

\begin{abstract}
	Time series prediction with deep learning methods, especially long short-term memory neural networks (LSTMs), have scored significant achievements in recent years. Despite the fact that the LSTMs can help to capture long-term dependencies, its ability to pay different degree of attention on sub-window feature within multiple time-steps is insufficient. To address this issue, an evolutionary attention-based LSTM training with competitive random search is proposed for multivariate time series prediction. By transferring shared parameters, an evolutionary attention learning approach is introduced to the LSTMs model. Thus, like that for biological evolution, the pattern for importance-based attention sampling can be confirmed during temporal relationship mining. To refrain from being trapped into partial optimization like traditional gradient-based methods, an evolutionary computation inspired competitive random search method is proposed, which can well configure the parameters in the attention layer. Experimental results have illustrated that the proposed model can achieve competetive prediction performance compared with other baseline methods.
\end{abstract}

\section{Introduction}
A time series is a series of data points indexed in time order. Effective prediction of time series can make better use of existing information for analysis and decision-making. Its wide range of applications includes but not limited to clinical medicine \cite{DBLP:conf/aaai/LiuSZWT18}, financial forecasting \cite{DBLP:conf/aaai/CaoHC15}, traffic flow prediction \cite{DBLP:conf/kdd/HulotAJ18}, human action prediction\cite{DBLP:conf/cvpr/DuWW15} and other fields. Different from other prediction modeling tasks, time series adds the complexity of sequence dependence among the input variables. It is crucial to build a suitable predictive model for the real data so as to make good use of the complex sequence dependencies.

The research on the time series prediction began with the introduction of regression equations \cite{Yule1927On} in the prediction of the number of sunspots over a year for the data analysis. The auto-regressive moving average model (ARMA) and auto-regressive integrated moving average model (ARIMA) \cite{Box1968Distribution} indicate that the time series prediction modeling based on the regression method gradually becomes mature. Therefore, such models also become the most basic and important ones in time series prediction. Due to the high complexity, irregularity, randomness and non-linearity of real data, it is very difficult for the methods above to achieve high-accuracy prediction through complex models. With machine learning methods, people build nonlinear prediction model based on a large number of historical time data. The fact is that we can obtain more accurate prediction results than traditional statistic-based models through repeated iterations of training and learning to approximate the real model. Typical methods such as support vector regression or classification \cite{DBLP:conf/nips/DruckerBKSV96} based on kernel method and artificial neural networks (ANN) \cite{DBLP:conf/dmin/DavoianL07} with the strong nonlinear function approximation ability and tree-based ensemble learning method, for instance, gradient boosting regression or decision tree (GBRT, GBDT) \cite{DBLP:conf/icmla/LiB16,DBLP:conf/nips/KeMFWCMYL17}. However, methods mentioned above begin to expose their own defects in dealing with the sequence dependence among input variables in time series prediction.

\begin{figure*}[t]
	\centering
	\includegraphics[width=0.95\linewidth,height=10.8cm]{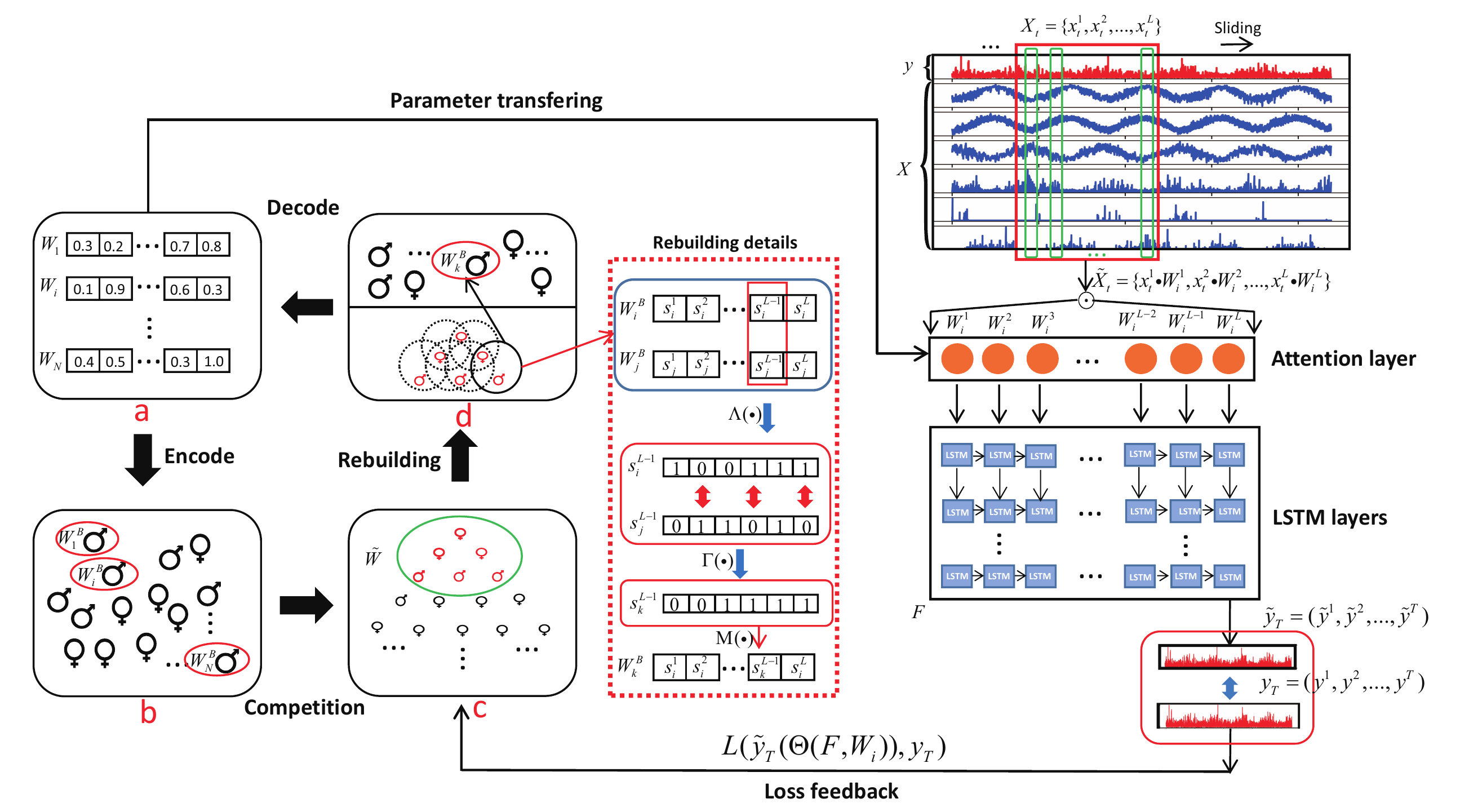}
	\caption{
		Graphical illustration of training evolutionary attention-based LSTM with competitive random search. This figure is composed of two parts. The left part displays the process of competitive random search, and the right part the structure of evolutionary attention-based LSTM. On the right, each sample: $X_t=(x_{t}^{1},x_{t}^{2},...,x_{t}^{L})$ in the training set $X=(X_1,X_2,...,X_T)$ multiplies attention weight $W_i$, the learning result of the left part, producing $\tilde{X_t}=(x_{t}^{1}W_{i}^{1},x_{t}^{2}W_{i}^{2},...,x_{t}^{L}W_{i}^{L})$, and $\tilde{X_t}$ are respectively sent to LSTMs for training. Finally, the error between the predication result $\tilde{y_T}$ and the real value $y_T$ is obtained in the validation set. The left part consists of a loop where the initial optimization sipace $W=(W_1,W_2,...,W_N)$ is established in ``a", and the subspace $W_i$ is encoded into $W^B=(W_{1}^{B},W_{2}^{B},...,W_{N}^{B})$ through binary code and sent to ``b". Meanwhile, $W_i$ are respectively transferred to the right network and the corresponding loss evaluation is gained in accordance with the prediction error of the network. Then, the champion subspace set $\tilde{W}$ is selected according to the loss situation of $\tilde{W}^{B}$ in ``c", and its subset combination is traversed repeatedly. Finally, the optimization space is reestablished in the light of operations in the red dotted box and $W$, the new-generation optimization space, is produced.
	}
	\label{f1}
\end{figure*}

The most commonly used and effective tool for time sequence model is recurrent neural network, or RNN \cite{Rumelhart1986Learning}. In normal neural networks, calculation results are mutually dependent, yet those of hidden layers in RNN are highly relevant to the current input as well as those produced last time in hidden layers. However, with longer driving sequence, problems such as vanishing gradient often appear in the training of RNN with commonly-used activation functions, \emph{e.g.}, tanh or sigmoid functions which limit the prediction accuracy of this model. The long short-term memory units (LSTMs) was proposed \cite{DBLP:journals/neco/HochreiterS97} based on the original RNN which mediates the balance between memorizing and forgetting by adding some multiple threshold gates. LSTMs and the gated recurrent unit (GRU) \cite{DBLP:conf/ssst/ChoMBB14} address the limited ability to deal with the long-term dependencies. These methods have led to the successful application for many sequence learning problems like machine translation \cite{DBLP:conf/emnlp/ChoMGBBSB14}. Therefore, the LSTMs is generally regarded as one of the state-of-the-art methods to deal with the time series prediction problem. Learning from cognitive neuroscience, some researchers introduce attention mechanisms to the encoding-decoding framework \cite{Bahdanau2014Neural} to better select from input series and encode the information in long-term memory to improve information processing ability. Recently, attention mechanisms have been widely used and performed well in many different types of deep learning tasks, such as image captioning \cite{DBLP:conf/cvpr/LuXPS17}, visual question answering \cite{DBLP:conf/iccv/YuY0T17} and speech recognition \cite{DBLP:conf/icassp/KimHW17}. Additionally, in recent years, some related work \cite{DBLP:conf/ijcai/QinSCCJC17,DBLP:conf/ijcai/LiangKZYZ18} on time series prediction is improved usually by introducing attention layers into the encoding-decoding framework.

Time series prediction is usually performed through sliding time-window feature and make prediction depends on the order of events. Firstly, we establish a multi-variate temporal prediction model based on LSTMs. Then, inspired by how human brain process input information with attention mechanism, we add an attention layer into the LSTMs. The introduced attention mechanism can quantitatively attach weight to period with diverse importance in the sliding time window so as to avoid being attention-distracted which is the primarily insufficient nature in traditional LSTMs. Specifically, instead of gradient-based methods, a competitive random search is employed to train the attention mechanism with reference to evolutionary computation and genetic algorithm\cite{DBLP:journals/siamcomp/Holland73}. When approximating the optimum solution, the random searching operators adopted can divert searching direction to the largest extent so as to avoid being trapped by local optimum solution\cite{2017arXiv171206564Z,2017arXiv171206560C,Lehman2017Safe}. As a result, compared with the traditional gradient-based one, competitive random search boasts stronger global searching ability when solving parameters in attention layer. So we can take advantage of this method to further improve the prediction accuracy of the LSTMs. To demonstrate the preformance, we conduct some experiments on real time series prediction datasets in both regression and classification tasks to compare it with some other baseline methods. The results show that the peoposed method can produce higher prediction accuracy than other baseline methods.

\section{Preliminaries}
In this section, formulation and description of the problem will be displayed. Time series prediction which can be divided into regression or classification problems usually uses a historical sequence of values as the input data. Given sliding-window feature matrix of training series $X=(X_1,X_2,...,X_T)$ and $X_t=(x_{t}^{1},x_{t}^{2},...,x_{t}^{L})$, where $X_t \in X $. Meanwhile, we define the length of time-step as $L$. Typically, historical values $y = (y_1,y_2,...,y_{T-1})$ are also given. As for classification problems, the historical values $y$ are discrete.

Generally, we learn a nonlinear mapping function by using the history-driven sequence feature $X$ and its corresponding target value $y$ to obtain the predicted value $\tilde{y}_{T}$ with the following formulation:
\begin{equation}
\label{e1}
\tilde{y_T}=f(X,y)
\end{equation}
where mapping $f(\cdot)$ is the nonliner mapping function we aim to learn.

\section{Methodology}
In this section, we will introduce the evolutionary attention based-LSTM and the competitive random search and present how to train this model in detail. In this part, we first give the overview of the model we proposed. Then, we will detail the evolutionary attention-based LSTM. Furthermore, we present the competitive random search and a collaborative training mechanism to train the model. A graphical illustration is shown in Figure \ref{f1}.

\subsection{Overview}
The idea of an evolutionary attention-based LSTM is to introduce a layer of attention to the basic LSTMs network. This enables the LSTMs networks not only to handle the long-term dependencies of drive sequences over historical time steps, but also an importance-based sampling. To avoid being trapped, we learn the attention weights by a competitive random search referring to evolutionary computation. To train the model, a collaborative meachaism is proposed. Attention weight that is learnt from the competitive random search is transferred to evolutionary attention-based LSTM networks for time series prediction. Meanwhile, predicted errors, as the feedback, are sent to direct the searching process. 

\subsection{Temporal Modeling with EA-LSTM}
Traditional methods generally model the time series prediction problem with hand-crafted features and make the prediction by well-designed regressors. Recurrent neural network (RNN) is chosen because of its capability to model long-term historical information of temporal sequences. Despite of so many basic LSTMs variants for capturing long-time dependencies proposed recently, a large-scale analysis shows that none of them can improve the performance in this issue significantly \cite{DBLP:journals/neco/HochreiterS97}. Therefore, we solve the problem of long-term dependence by replacing the simple RNN unit with the LSTMs neuron structure in the recurrent neural network. The LSTMs is a special kind of RNN. With its gated structure, including the forget gate, the input gate and the output gate, LSTMs can memorize what should be memorized and forget what should be forgot. Especially, the forget gate is the first operator in LSTMs to decide what information in last time-step should be dropped with a sigmoid function. It is a key operator in its gated structure. 

Firstly, we define the attention weights as:
\begin{equation}
W=(W^1,W^2,...,W^L)
\end{equation}
with these attention weights, we can take importance-based sampling for input data with
\begin{equation}
\tilde{X_t}=(x_{t}^{1}W^{1},x_{t}^{2}W^{2},...,x_{t}^{L}W^{L})
\end{equation}
Then, $\tilde{X}=(\tilde{X_1},\tilde{X_2},...,\tilde{X_T})$ is fed into LSTM networks. Furthermore, we can learn the nonliner mapping function by these formulations \cite{DBLP:series/sci/2012-385} of the calculating process in LSTMs cells as follows:
\begin{equation}
i^t=\sigma(W_{xi}\tilde{X_t}+W_{hi}h^{t-1}+W_{ci}c^{t-1}+b_i)
\end{equation}
\begin{equation}
f^t=\sigma(W_{xf}\tilde{X_t}+W_{hf}h^{t-1}+W_{cf}c^{t-1}+b_f)
\end{equation}
\begin{equation}
c^t=f^tc^{t-1}+i^t\tanh(W_{xc}\tilde{X_t}+W_{hc}h^{t-1}+b_c)
\end{equation}
\begin{equation}
o^t=\sigma(W_{xo}\tilde{X_t}+W_{ho}h^{t-1}+W_{co}c^{t-1}+b_o)
\end{equation}
\begin{equation}
h^t=o^t\tanh(c^t)
\end{equation}
where $\sigma(\cdot)$ represents the activation function of sigmoid and $W$ matrices with double subscript the connection weights between the two cells. In addition, $i^t$ represents input gate state, $f^t$ forget gate state, $c^t$ cell state, $o^t$ output gate and $h^t$ the hidden layer output in current time-step. Finally, we can take the last element of output vector $h^{t-1}$ as the predicted value. It can be represented as:
\begin{equation}
\tilde{y}^{t}=h^{t-1}
\end{equation}
the final output value can be contacted to a vector:
\begin{equation}
\tilde{y_T}=(\tilde{y}^{1},\tilde{y}^{2},...,\tilde{y}^{T})
\end{equation}

\subsection{Competitive Random Search}
\renewcommand{\algorithmicrequire}{ \textbf{Input:}} %Use  Input in the format of Algorithm
\renewcommand{\algorithmicensure}{ \textbf{Output:}} %Use Output in the format of

Based on genetic algorithm, competitive random search (CRS) is proposed to generate the optimum parameter combinations in the attention layer of LSTM network. The detailed process of the CRS is elaborated in Figure \ref{f1}. The CRS consists of four parts which are introduced as follows.  

In Figure \ref{f1}, attention weights set $W=(W_1,W_2,...,W_N)$ is given in ``a". While being translated into $W^B=(W_{1}^{B},W_{2}^{B},...,W_{N}^{B})$ through binary code and sent into ``b", the subset $W_i$ which denotes attention weights are transferred into networks in the right part and produce a corresponding loss value according to predicted error in the networks. Then, the champion attention weights subset $\tilde{W}$ is selected according to the loss of $W^B=(W_{1}^{B},W_{2}^{B},...,W_{N}^{B})$ in ``c", and its subset combination is traversed repeatedly. Finally, as is shown in the red dotted box, a new attention wights is rebuilt and $W_{k}^{B}$, the new-generation optimization subspace, is produced. 

Random operators are introduced to illustrate how optimization space is rebuilt in ``d". In the red dotted box in Figure \ref{f1}, if the selected champion combination is $W_{i}^{B}$ and $W_{j}^{B}$ where each individual is composed of binary strings, they will be evenly divided into $L$ segments in line with L, the time step defined in section 2.1. Then, the corresponding $W_{i}^{B}$ can be expressed by $W_{i}^{B}=(S_{i}^{1},S_{i}^{2},...,S_{i}^{L})$ where $S_{i}^{1}$ is a segment of $W_{i}^{B}$. Two important operators are described as follows:
\begin{itemize}
	\item {
		\textbf{Randomly select.} Firstly, we define this opreator as $\Lambda(\cdot)$. Its function is to randomly select subsegments in each champion combination. For instance, in Figure \ref{f1}, the subsegment $L-1$ of the two subspaces are selected. It should be noted that the number of selected subsection is not fixed. 
	}
	\item {
		\textbf{Recombine.} This opreator can be expressed as $\Gamma(\cdot)$. It is defined to recombine the genes in the selected subsegment. The process interchanges the two subsegments expressed by binary codes with the length of 6 and from different subspaces in either even or odd index. $s_{i}^{L-1}$ and $s_{j}^{L-1}$ will generate $s_{k}^{L-1}$ after $\Gamma(\cdot)$. It should also be noted that the figure only displays interchange in the even index, but the index where actual interchanges happen is decided by the random judgment of $\Gamma(\cdot)$.
	}
\end{itemize}
After the abovementioned two steps, gene mutation, a link in biological evolution, is imitated. The operator $M(\cdot)$ is set to reverse the genotype of the newly generated $s_{k}^{L-1}$ in a random index. For instance, 0 is reversed to 1. Finally, $s_{k}^{L-1}$ replaces the corresponding $s_{i}^{L-1}$ in $W_{i}^{B}$, forming $W_{k}^{B}$ which is inserted into $W$. When rebuilding optimization space, we will repeatedly traverse subspace $\tilde{W}$ until the size of $W$ has reached the default value $N$. The key factor in the CRS is the error feedback introduced from the right network in Figure \ref{f1}. The CRS is demonstrated in the optimizing issue as follows:

\begin{algorithm}[t]
	\caption{Competitive Random Search}
	\label{active_sample}
	\begin{algorithmic}[1]
		\REQUIRE ~~\\
		$N$: size of attention weights set, $T$: epochs , $\tilde{W}$: champion attention weights subset with the size of $\tilde{N}$, $L=(L_1,L_2,...,L_N)$: loss of each $W_i \in W$
		\ENSURE ~~ \\
		$W$: attention weights set
		\WHILE{$t<T$}
		\IF {$t=0$}
		\STATE $W\leftarrow$$(W_1,W_2,...,W_N)$
		\ELSE
		\STATE $\tilde{W}\leftarrow$$Ranking(W_i|L_i,\tilde{N})$
		\STATE $W\leftarrow\emptyset$
		\WHILE{$length(W)\textless N$}
		\STATE $W\leftarrow\tilde{W}$
		\FOR{$(W_i,W_j)\in \tilde{W}$}
		\STATE $W_k\leftarrow M(\Gamma(\Lambda(W_i,W_j)))$
		\ENDFOR
		\STATE $W\leftarrow W_k$
		\ENDWHILE 
		\ENDIF
		\ENDWHILE
	\end{algorithmic}
\end{algorithm}

\begin{equation}
\min L( \tilde{y_T}(\Theta(F,W)),y_T)
\end{equation} 
where $F$ are entire parameters in LSTM networks and $\Theta(\cdot)$ is the parameter space needed when obtaining the predicted value $\tilde{y_T}$. 
The most important operator is the rebuilding process which can determines the performance significantly by controlling the direction of the random searching. Algoithm 1 outlines the competitive random search. 

\subsection{Parameters Transferring}
To train our model, we proposed a collaborative mechanism which combines EA-LSTM with the competitive random search. The idea of collaborative training is to share the parameters and loss feedback between the two components of the model. We use mini-batch stochastic gradient decent (SGD) together with Adam optimizer \cite{DBLP:journals/corr/KingmaB14} to train EA-LSTM. Except for attention layer, the other parameter in LSTMs can be learned by standard back propagation through time algorithm with mean squared error and cross entropy loss as the objective function. Meanwhile, the attention weights outputted by competitive random search will be fed into attention layer before the LSTM network begin to be trained. In addition, the current prediction loss of the LSTMs in validation set will be used to rank the optimization space.

\section{Experiments}
In this section, the description of datasets used in our research is given firstly. Then we will introduce the parameter settings and show the training result of EA-LSTM. Furthermore, we compare the model we proposed with some baseline models \emph{e.g.}, SVR, GBRT, RNN, GRU and LSTM. In addition, several attention-based methods also as the competitors to verify the performance of our proposed model.

\subsection{Datasets Description and Setup }
To compare the performance of different models with varied types of time series prediction problem, datasets used in our experiments are described as follows: 

\begin{figure*}[t]
	\centering
	\subfigure{
		\includegraphics[width=0.28\linewidth,height=4.5cm]{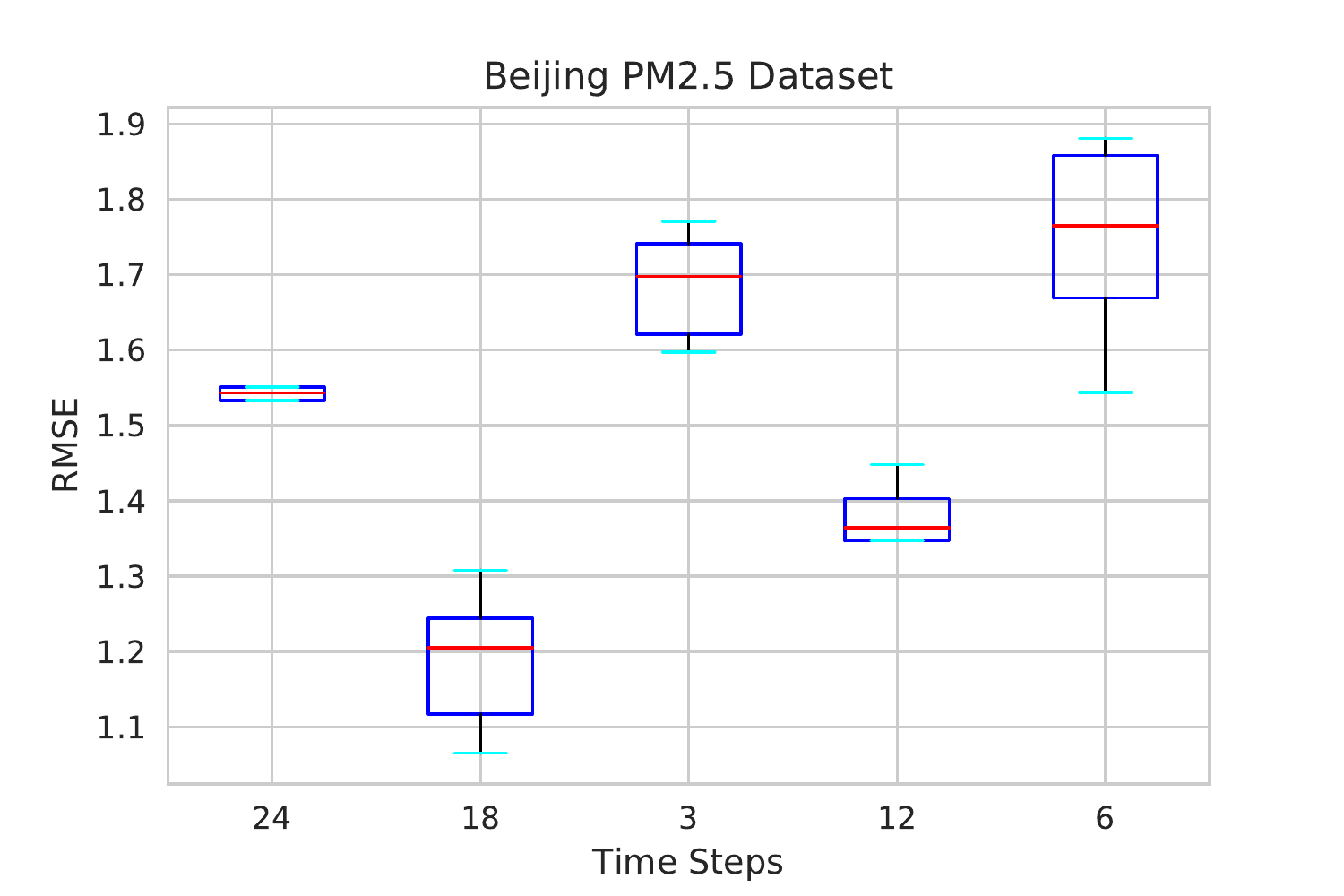}
	}
	\subfigure{
		\includegraphics[width=0.28\linewidth,height=4.5cm]{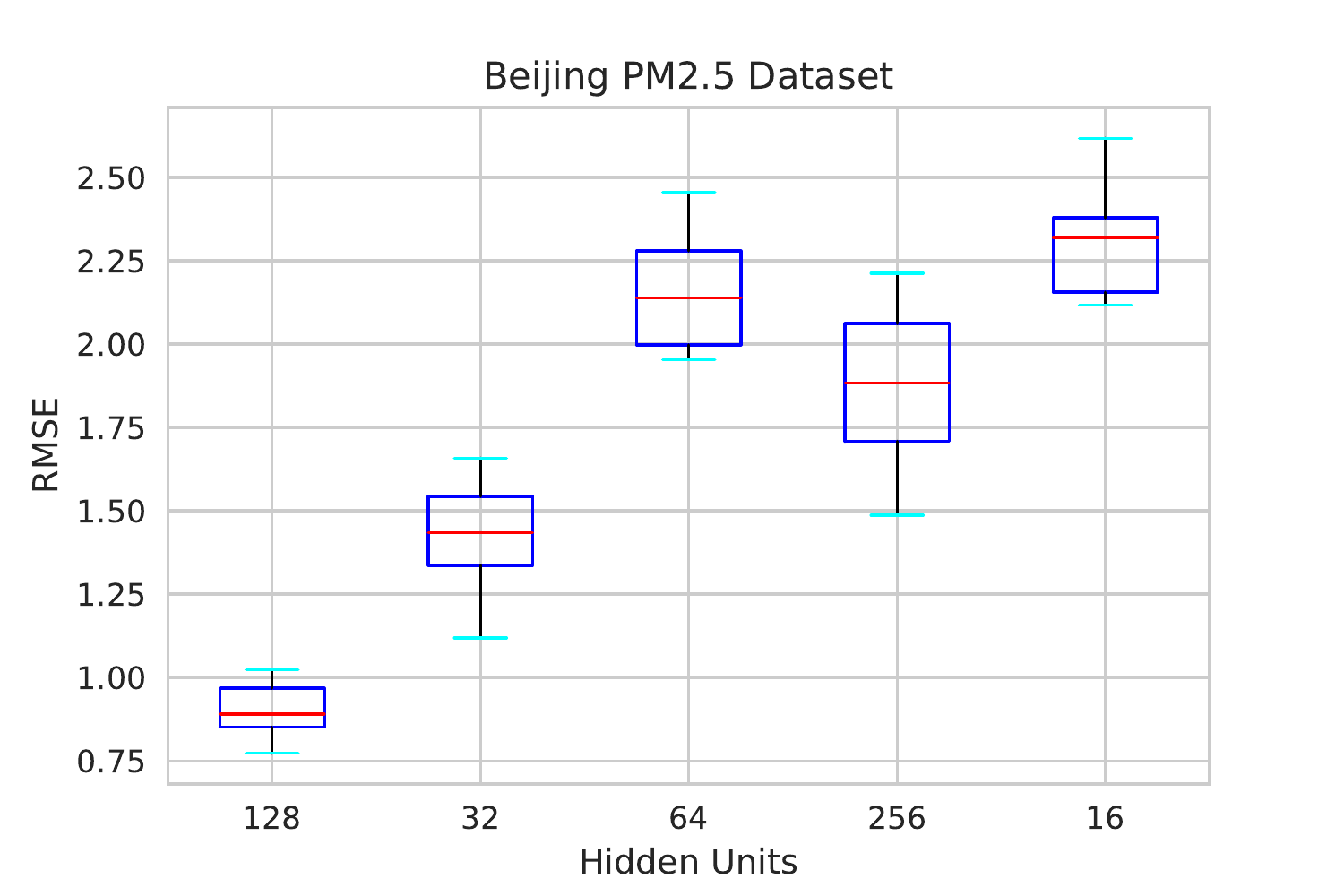}
	}
	\subfigure{
		\includegraphics[width=0.28\linewidth,height=4.5cm]{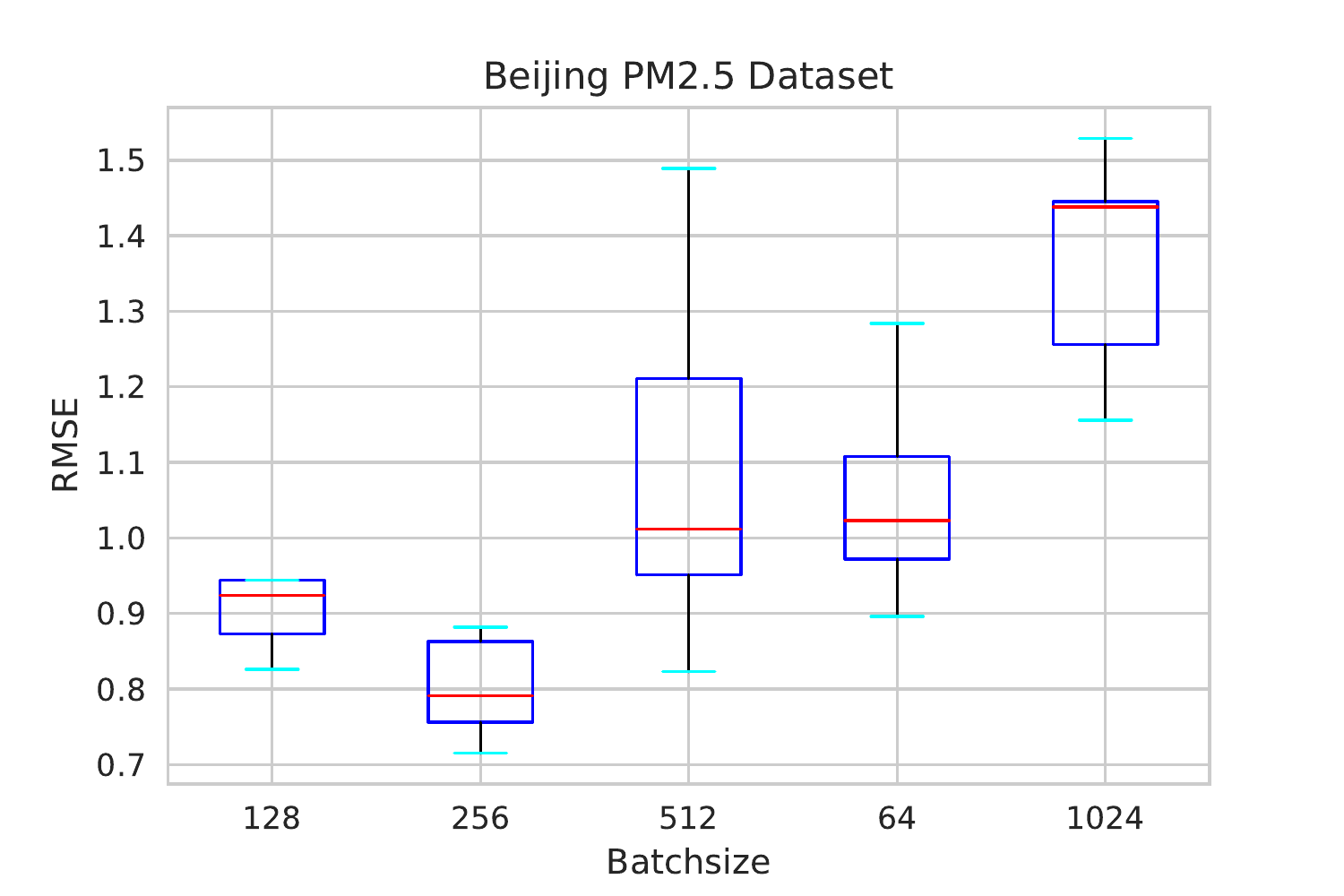}
	}
	\caption{Parameter Sensitivity in Beijing PM2.5 Dataset}
	\label{f2}
\end{figure*}
\begin{figure*}[t]
	\centering
	\subfigure{
		\includegraphics[width=0.28\linewidth,height=4.5cm]{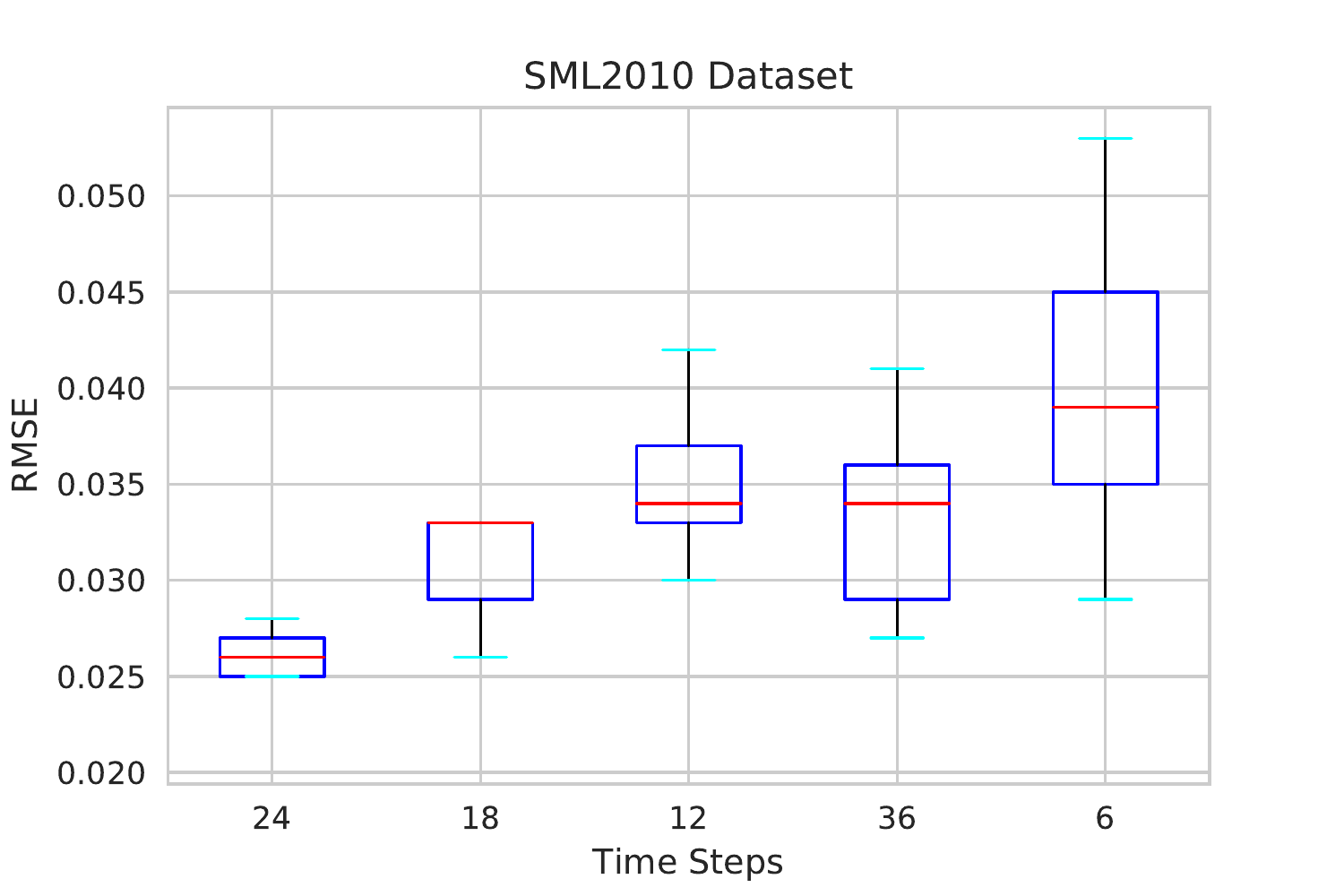}
	}
	\subfigure{
		\includegraphics[width=0.28\linewidth,height=4.5cm]{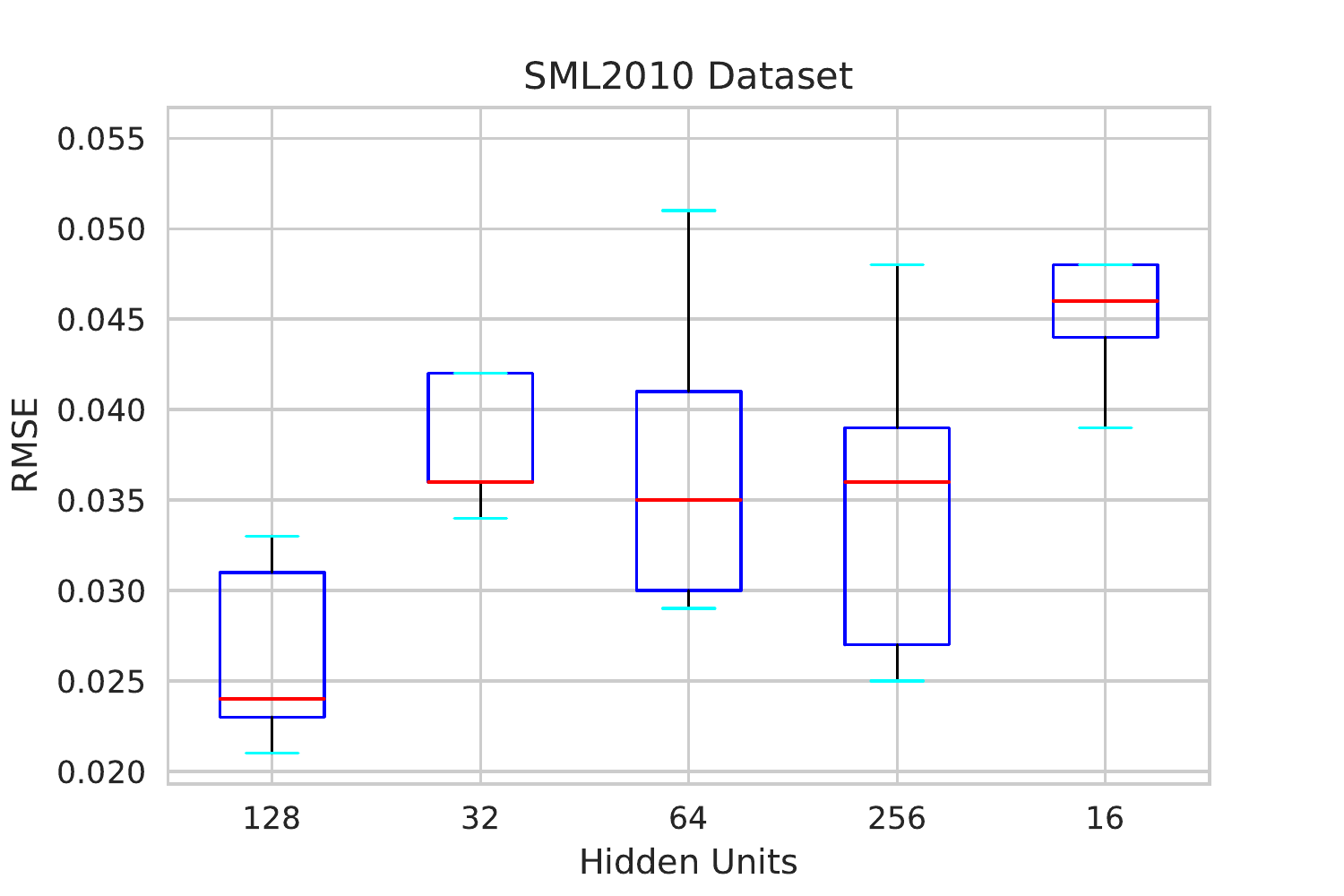}
	}
	\subfigure{
		\includegraphics[width=0.28\linewidth,height=4.5cm]{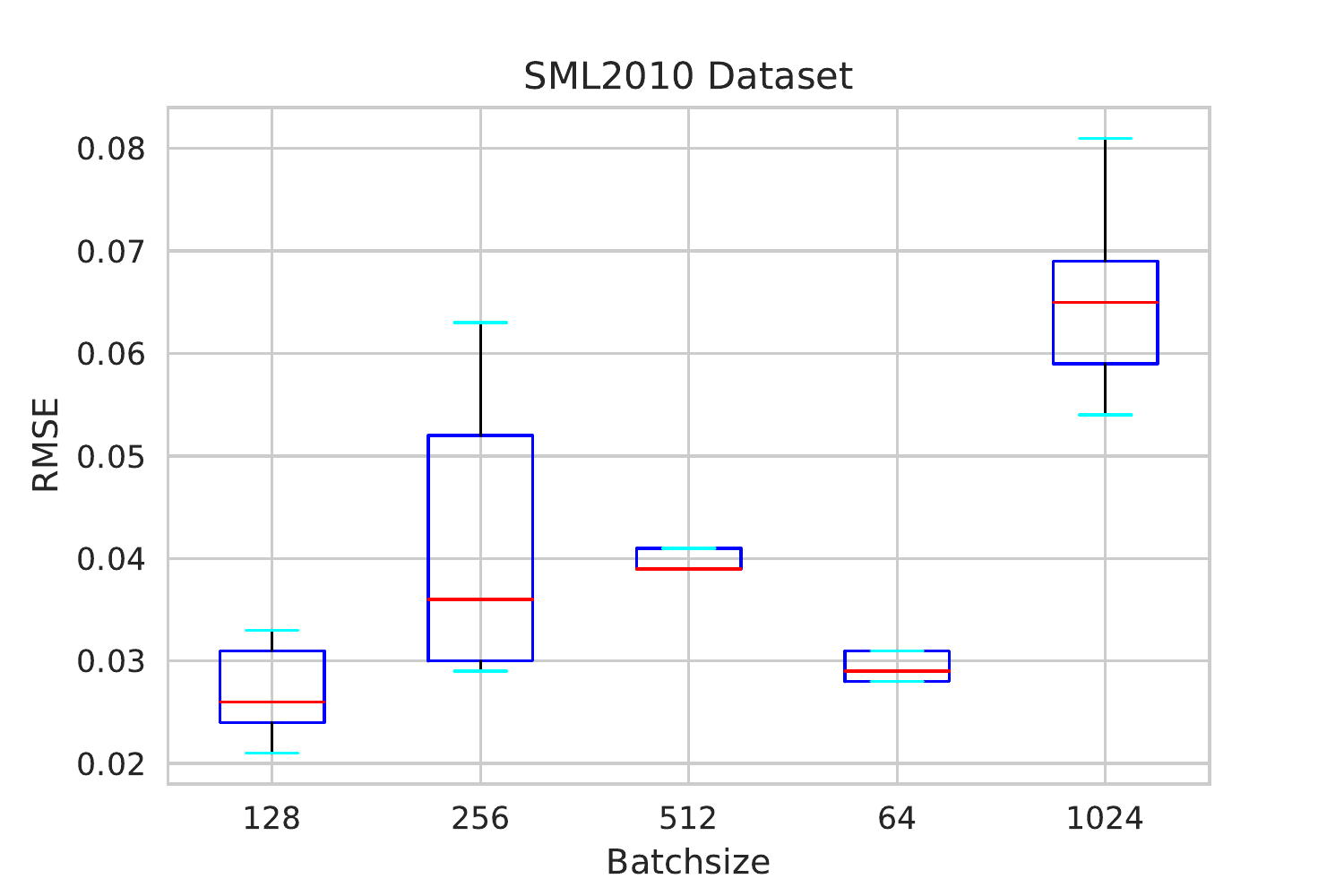}
	}
	\caption{Parameter Sensitivity in SML2010 Dataset}
	\label{f3}
\end{figure*}

\begin{itemize}
	\item {
		\textbf{Beijing PM2.5 Data\footnote{http://archive.ics.uci.edu/ml/datasets/Beijing+PM2.5+Data}} (PM2.5). This dataset \cite{Liang2015Assessing} contains the PM2.5 data of US Embassy in Beijing with an hour sampling rate between January 1st, 2010 and December 31st, 2014. Meanwhile, meteorological data from Beijing Capital International Airport are also included. Its sensor data \emph{e.g.}, current time, PM2.5 concentration, dew point, temperature, pressure, wind direction, wind speed, hours of snow, hours of rain. The PM2.5 concentration is the target value to predict in the experiments.
	}
	\item {
		\textbf{SML2010\footnote{http://archive.ics.uci.edu/ml/datasets/SML2010}} (SML). It is a uci open dataset \cite{zamora2014line} used for indoor temperature prediction. This dataset is collected from a monitor system mounted in a domotic house. It corresponds to approximately 40 days of monitoring data. The data was sampled every minute, computing and uploading it smoothed with 15 minute means. The sensor data we use includes current time, weather forecast temperature, carbon dioxide, relative humidity, lighting, rain, sun dusk, wind, sun light in west facade, sun light in east facade, sun light in south facade, sun irradiance, Enthalpic motor 1 and 2, Enthalpic motor turbo, outdoor temperature, outdoor relative humidity, and day of the week. The room temperature is the target value to predict in our experiments.
	}
	\item{
		\textbf{MSR Action3D Dataset\footnote{http://research.microsoft.com/en-us/um/people/zliu/actionrecorsrc/.}} (MSR). MSR Action3D dataset contains twenty actions: high arm wave, horizontal arm wave, hammer, hand catch, forward punch, high throw, draw x, draw tick, draw circle, hand clap, two hand wave, side-boxing, bend, forward kick, side kick, jogging, tennis swing, tennis serve, golf swing, pick up and throw. There are 10 subjects, each subject performs each action 2 or 3 times. There are 567 depth map sequences in total. The resolution is 320x240. The data was recorded with a depth sensor similar to the Kinect device.
	}
\end{itemize}

\begin{figure*}[t]
	\centering
	\subfigure{
		\includegraphics[width=0.3\linewidth,height=5.3cm]{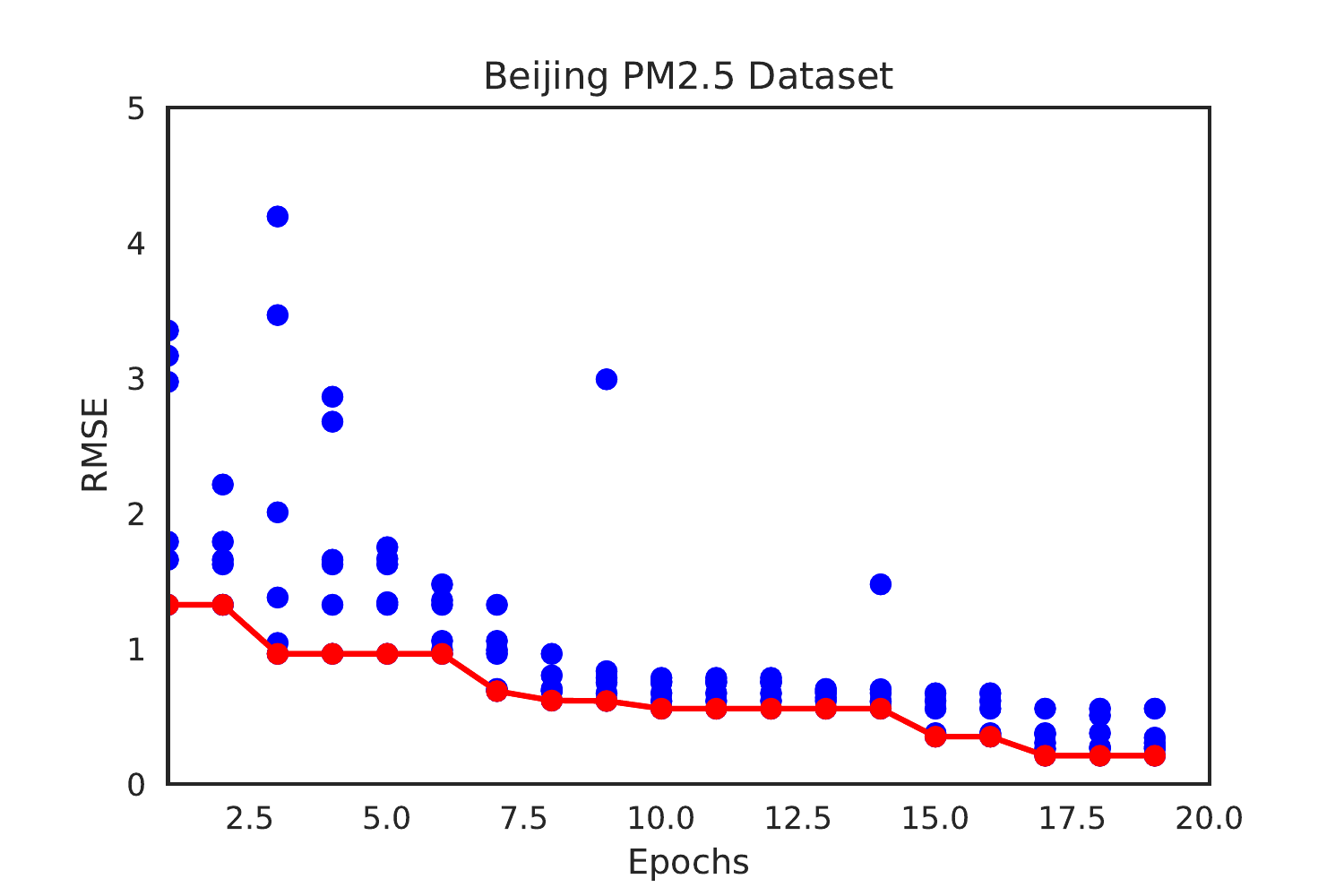}
	}
	\subfigure{
		\includegraphics[width=0.3\linewidth,height=5.3cm]{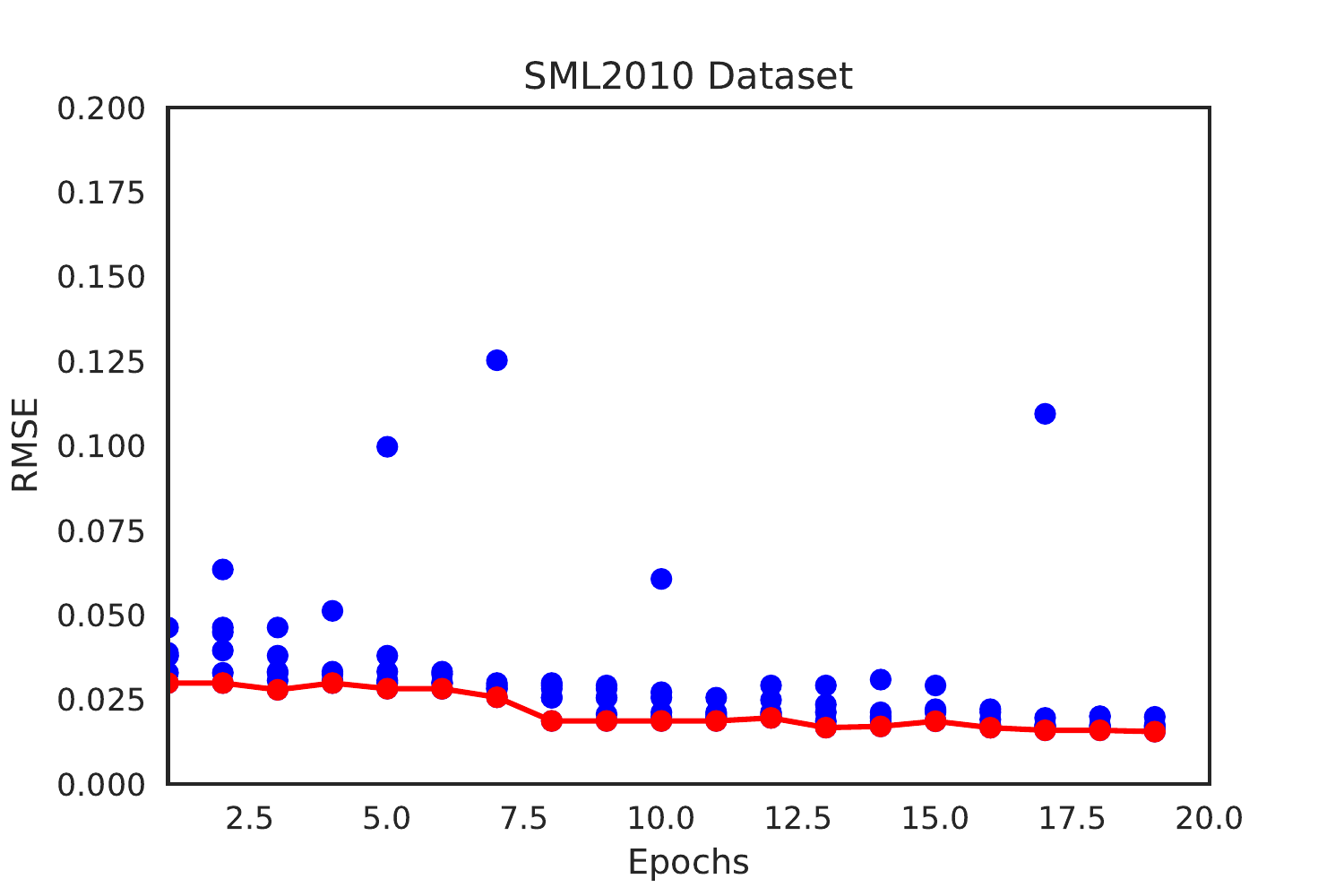}
	}
	\subfigure{
		\includegraphics[width=0.3\linewidth,height=5.3cm]{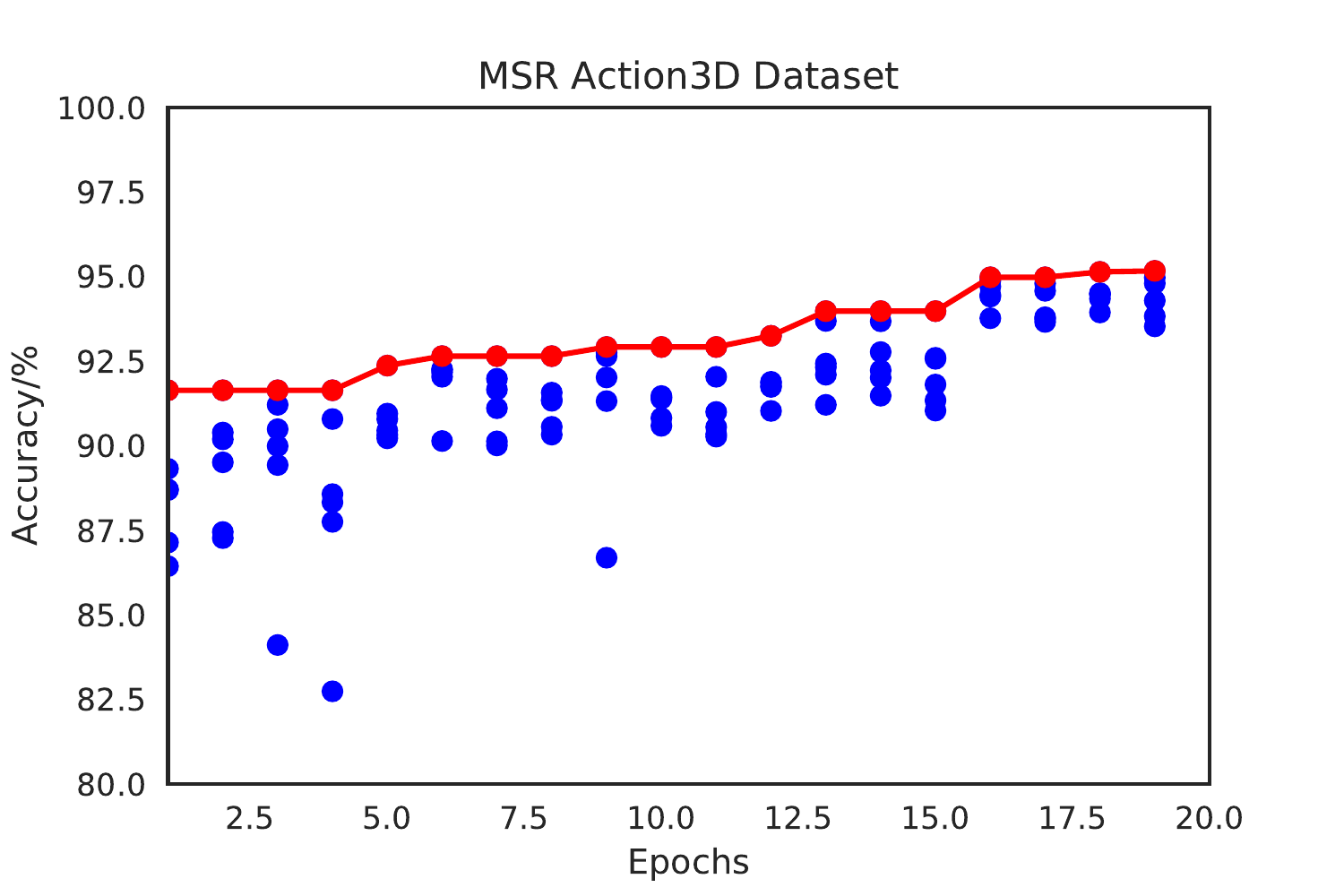}
	}
	\caption{Training Process of Competitive Random Search. The points drawn in figure indicates the error calculation results of each weight in champion subspace $\tilde{W}$ for each epoch. There are six champion weights with blue spot and the best one with red spot for each epoch in the figure. It should be noted that points in the third subfigure shows the accuracy curve of EA-LSTM.}
	\label{f4}
\end{figure*}

The setting for PM2.5 and SML dataset is given in Table \ref{t1}. In addition, as for MSR dataset, we follow the standard setting provided in \cite{DBLP:conf/cvpr/DuWW15} and calculate the average accuracy for comparison.

\begin{table}[t]
	\centering
	\caption{Statistic of Two Datasets for Regression Tasks}
	\begin{tabular}{cccc}
		\hline
		\hline
		Dataset&Sensors&Train \& Vaild&Test\\
		\hline
		Beijing PM2.5&	  8&   35,040&  8,760\\
		SML 2010&      16&  3,600&	537\\ 
		\hline
		\hline
	\end{tabular}
	\label{t1}
\end{table}

\subsection{Parameter Settings and Sensitivity}
There are three parameters in the basic LSTM model, \emph{i.e.}, the number of time steps $L$ and the size of hidden units for each layers in LSTM $m$ (we set the same hidden units for each layer in LSTM) and the batchsize $b$ in training process. We carefully tuned the parameters $L$ (time-steps), $m$ (hidden units number) and $b$ (batchsize) for our basic model. To approximate the best performance of the model, we conducted a grid search over $L \in \{3,6,12,18,24\}$, $m \in \{16,32,64,128,256\}$, $b \in  \{64,128,256,512,1024\}$ in the \textbf{Beijing PM2.5} dataset and $L \in \{6,12,18,24,36\}$ ,  $m \in \{16,32,64,128,256\}$, $b \in \{64,128,256,512,1024\}$ in the \textbf{SML2010} dataset and $m\in \{16,32,64,128\}$, $b \in\{8,16,32,64\}$ in the \textbf{MSR Action3D} dataset. It should be noted that in MSR dataset we set the number of frames in each sample as the $L$. When one parameter vaies, the others are fixed. Finally, we achieve the hyperparameters with the best performance over the validation set which are used to fix the basic model structure. The box line diagram plotted in Figure \ref{f2}-\ref{f3} is used to show the sensitivity of the parameters on two dataset used for regression tasks. 

The root means squared error for the time series task with one box-whisker (showing middle value, 25\% and 75\% quantiles, minimum, maximum and outliers) for five testing results of the basic model we proposed. After grid searching, we define the hyperparameters used in EA-LSTM with the best ones. The hyperparameters of LSTM in differents datasets are given in Table \ref{t2}.

Furthermore, there are four hyperparameters in competitive random search, \emph{i.e.}, the size of attention weights set $N$, the encoding length for each attention weights, the size of champion attention weights subset $\tilde{W}$ and the number of epochs $T$. To balance the solving efficiency, we defined size of optimization apace as 36, encoding length for each subspace as 6 which varies from 0.016 to 1.000, the size of $\tilde{W}$ as 6, and the number of epochs $T$ as 20.

\begin{table}[h]
	\centering
	\caption{Hyperparameters of LSTM in Each Dataset}
	\begin{tabular}{cccc}
		\hline
		\hline
		Dataset&Time-Steps&Units&Batchsize\\
		\hline
		Beijing PM2.5&  18	&128	&256\\
		SML 2010&	24	&128    &128\\
		MSR Action3D&    13  &128    &16\\
		\hline
		\hline
	\end{tabular}
	\label{t2}
\end{table}

\subsection{Evaluation Metrics}
To evaluate the performance, we take the Root Mean Squared Errors (RMSE) \cite{DBLP:journals/nn/PlutowskiCW96} and Mean Absolute Errors (MAE) as the evaluation metrics. They are calculated by the following.
\begin{equation}
RMSE=\sqrt{\frac{1}{N}\sum_{i=1}^{N}(\tilde{y}_{t}^{i}-y_{t}^{i})^2}
\end{equation}
\begin{equation}
MAE=\frac{1}{N}\sum_{i=1}^{N}|\tilde{y}_{t}^{i}-y_{t}^{i}|
\end{equation}
where $\tilde{y}_{t}^{i}$ is prediction, $y_{t}^{i}$ is real value and $N$ is the number of testing samples.

\subsection{Training Attention Layer}

\begin{table*}[htbp]
	\centering
	\caption{Performance of Different Baseline Methods Compared in Two Datasets for Regression Tasks}
	\begin{tabular}{|p{5cm}<{\centering}|p{2.5cm}<{\centering}|p{2.5cm}<{\centering}|p{2.5cm}<{\centering}|p{2.5cm}<{\centering}|}
		\hline
		\hline
		\multirow{3}{*}{Model}&
		\multicolumn{4}{c|}{Datasets}\\
		\cline{2-5}
		&\multicolumn{2}{c|}{Beijing PM2.5}
		&\multicolumn{2}{c|}{SML2010} \\
		
		\cline{2-5}
		&MAE	&RMSE	&MAE	&RMSE\\
		\cline{1-5}
		\hline
		SVR&   2.6779	&2.8623    &0.0558&	 0.0652\\
		GBRT&	0.9909	&1.0576    &0.0253&	 0.0327\\
		RNN&	0.8646  &0.9621    &0.0261&	 0.0367\\
		GRU&	0.6733	&0.7433	   &0.0231&	 0.0288\\
		LSTM&	0.6168  &0.7026    &0.0178&	 0.0234\\
		Attention-LSTM&	0.2324&	0.3619&	0.0190&	0.0225\\
		DA-RNN \cite{DBLP:conf/ijcai/QinSCCJC17}&	——	&——&	0.0150	&0.0197\\
		EA-LSTM&	\textbf{0.1902}&	\textbf{0.2755}&	\textbf{0.0103}&	\textbf{0.0154}\\
		\hline
		\hline
	\end{tabular}
	\label{t3}
\end{table*}

\begin{figure}[h]
	\centering
	\subfigure{
		\includegraphics[width=0.45\linewidth,height=2.5cm]{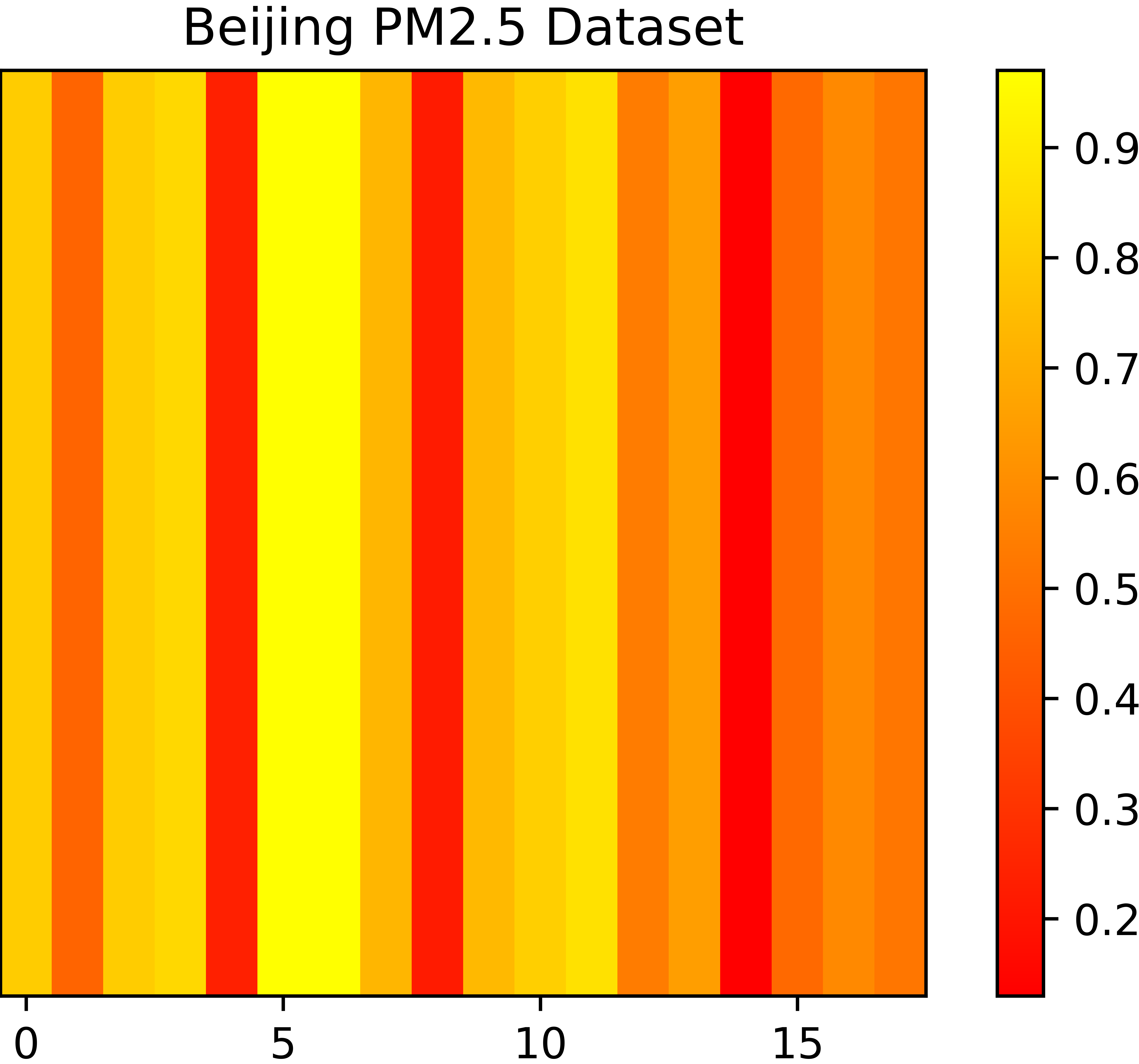}
	}
	\subfigure{
		\includegraphics[width=0.45\linewidth,height=2.5cm]{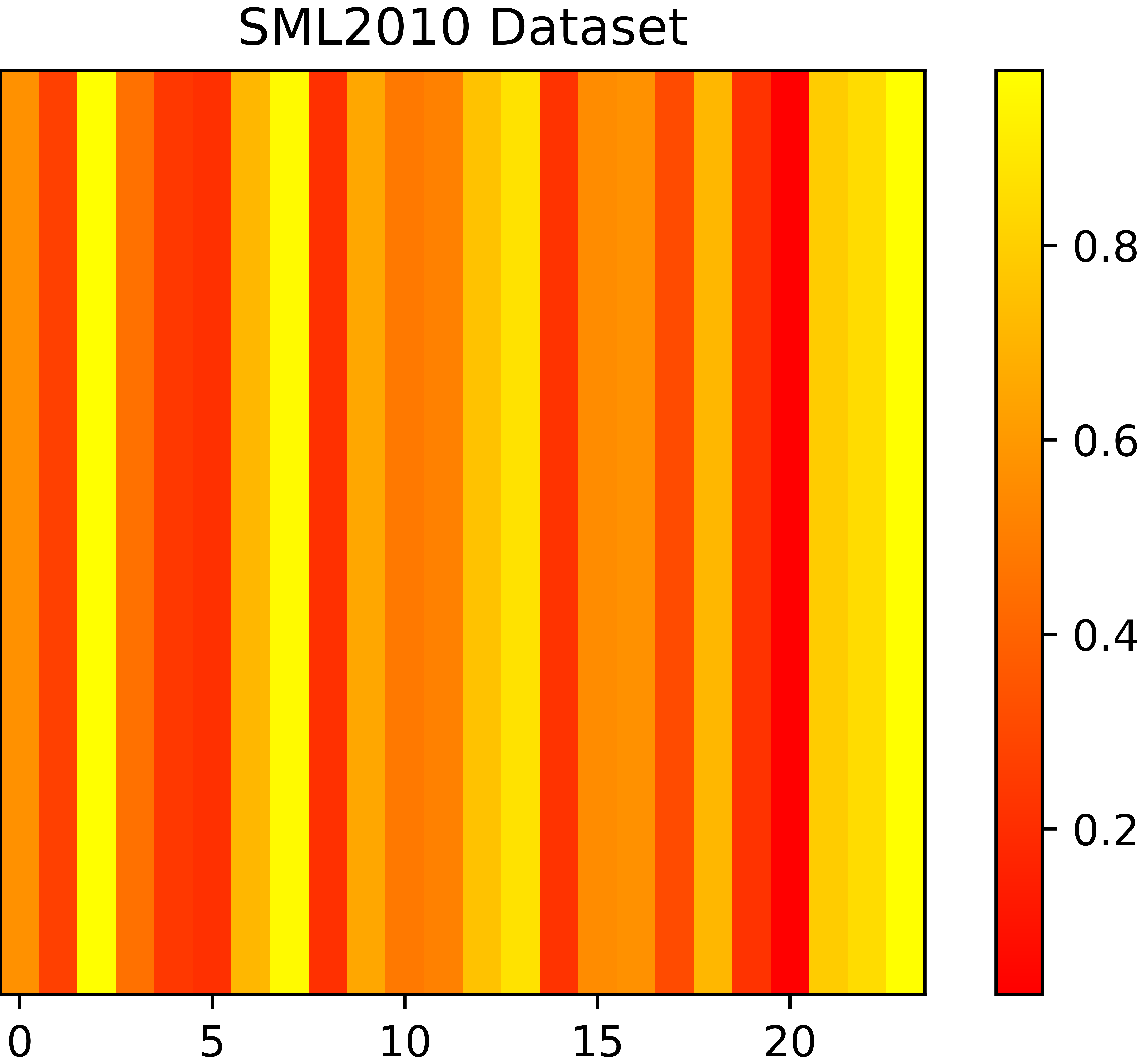}
	}
	\subfigure{
		\includegraphics[width=0.45\linewidth,height=2.5cm]{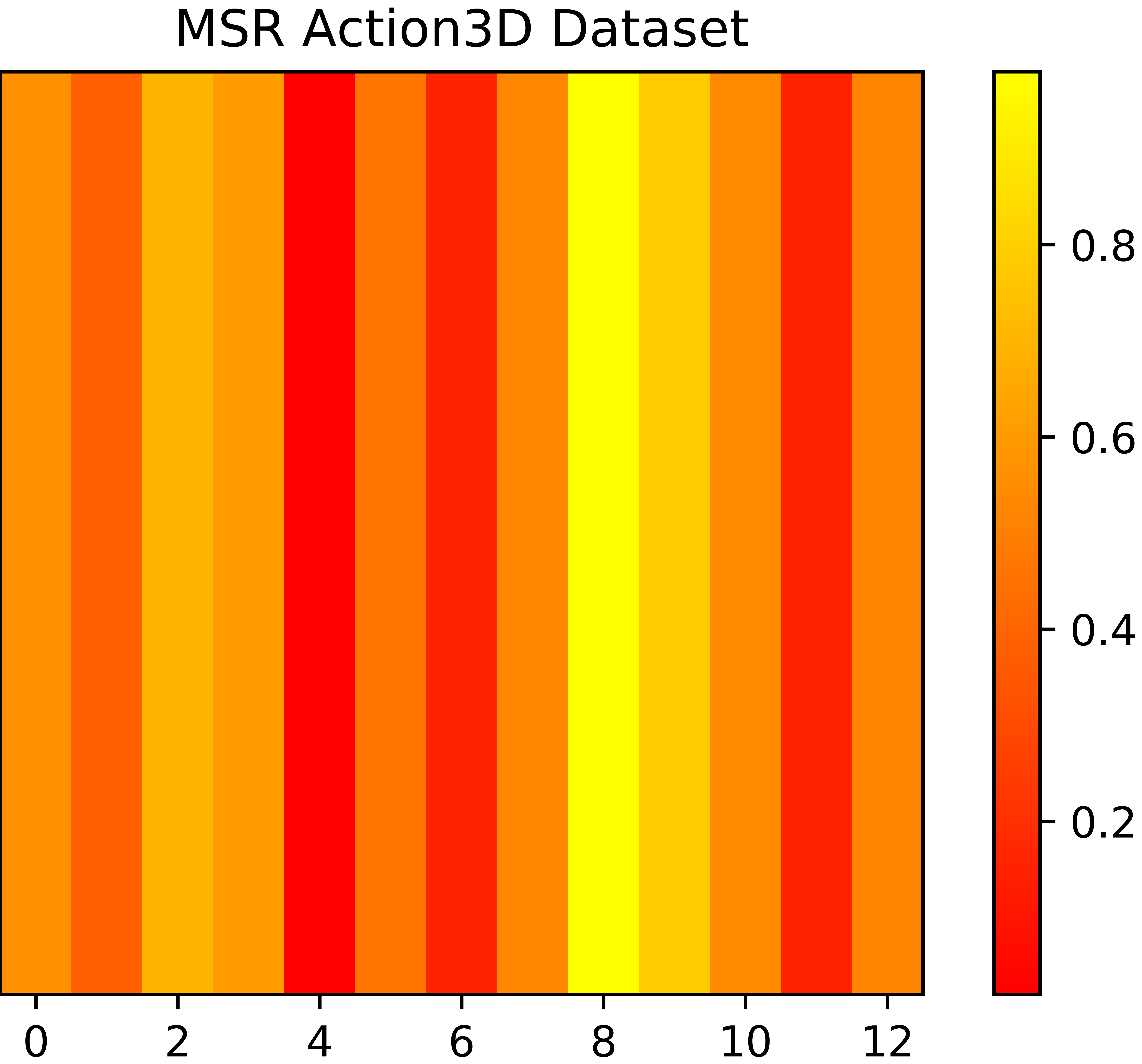}
	}
	\caption{Plot of the attention distribution for each time-step of the features extracted by sliding-time-window in three experimental datasets, in which the coordinates represent the time steps of the input driving series.}
	\label{f5}
\end{figure}

We trained the EA-LSTM with Competitive random search for 20 epochs. Training processes are visualized in Figure \ref{f4}. In addition, the points drawn in Figure \ref{f4} indicate the error and accuracy corresponding to the weights selected in champion subspace $\tilde{W}$. We can find that training with the CRS, attention weights in optimization space continuously improve the performance and not be trapped. Meanwhile, to better understand importance-based sampling of input series within time steps, the most suitable attention weights are visualized by heat map and showed in Figure\ref{f5}. In Figure \ref{f5}, varied scale of attention distribution of input driving series within multiple time steps over each datasets are showed as well. By solving attention weights which can better suits for the characteristics across different tasks, we improve the performance of the LSTMs and get better prediction results. In addition, we can also find that the proposed method effectively utilize local information within one sampling window according to varied scale of attention distribution in Figure \ref{f5}. It is crucial to make a soft feature selection in multiple time steps time series prediction.

\subsection{Performance Comparison}
To evaluate the performance of the EA-LSTM training with competitive random search in time series prediction, we set contrast experiments with some other baseline methods, including traditional machine learning methods and deep learning methods. In experiments, the SVR, GBRT, RNN, LSTMs and GRU as competitors are carefully tuned respectively. In addition, all the baseline methods we compared were trained and tested for five times and final prediction results showed in Table \ref{t3} were averaged to reduce random errors. We can see that the proposed method effectively improved the performance against to its baseline counterparts in both public open benchmarking datasets uaually uesd for time series prediction.

Furthermore, we also compared the proposed method with DA-RNN \cite{DBLP:conf/ijcai/QinSCCJC17} in our public testing dataset: \textbf{SML 2010}. DA-RNN, which is similar to the traditional attention-based model, is a time series predictive model trained by solving the network parameters together with attention-layer parameters. As a matter of fact, this model obtained the state-of-the-art performance by constructing a more complex attention mechanism. With the dataset identically classified into sets for training, validating, and testing, the experimental results show that the EA-LSTM can get a higher predicted precision. We can also see that there is the feasibility to enhance attention-based model by improving training method for attention layer not only by introducing a more complex attention mechanism. 

In addition, we compared the proposed method with the same method whose optimization method is replaced with gradient descent which named "Attention-LSTM" to clearly highlight the benefit of using the competitive random search, instead of gradient descent. Specifically, an input-attention layer whose weights are learned together with other parameters is introduced to LSTM networks. The experimental results clearly highlight the benefit of using the evolutionary computation inspired competitive random search to refrain from being trapped into partial optimization effectively, instead of gradient descent. 

\begin{table}[H] %开始一个表格environment，表格的位置是h,here。  
	\centering
	\caption{Experimental Results on The MSR Action3D Dataset.} %显示表格的标题  
	\begin{tabular}{p{6cm}<{\centering}|p{1.5cm}<{\centering}} %设置了每一列的宽度，强制转换。  
		\hline  
		\hline  
		Methods & Accuracy/\% \\ %用&来分隔单元格的内容 \\表示进入下一行  
		\hline %画一个横线，下面的就都是一样了，这里一共有4行内容  
		\cite{DBLP:conf/ijcai/GowayyedTHE13}  & 91.26\\
		\hline 
		\cite{DBLP:conf/cvpr/VemulapalliAC14}  & 92.46\\
		\hline
		HBRNN \cite{DBLP:conf/cvpr/DuWW15}  & 94.49\\
		\hline
		LSTM & 90.67 \\  
		\hline  
		Attention-LSTM  & 92.58 \\
		\hline
		EA-LSTM  & \textbf{95.20}\\  
		\hline  
		\hline  
	\end{tabular} 
	\label{t4} 
\end{table}

More general, we also add a comparison between our proposed model and some baseline ones by human action recognition experiments, a typical time series prediction task for testing the ability to take temporal modeling of different methods. The experimental results testify that the proposed method also delivers robust performance even in classification prediction tasks.

\section{Conclusion}
This paper proposed an evolutionary attention-based LSTM model (EA-LSTM) which is trained with competitive random search for time series prediction. The parameters of attention layer used for importance-based sampling in the proposed EA-LSTM networks can be confirmed during temporal relationship mining. Thus, this network is able to properly settle the local feature relationship within time-steps. As a hard optimization issue to approximate the best attention weights for real input driving series data, the CRS method we proposed can avoid being trapped during the parameters solving. Experiments show the EA-LSTM can make competitive prediction performance compared with the state-of-the-art methods. These results demonstrate that training evolutionary attention-based LSTM with competitive random search can not only help to capture long-term dependencies in time series prediction, but also effectively utilize local information within one sampling window according to varied scale of attention distribution. Besides, taking genetic algorithm as an example, this paper introduces evolutionary computation to substructure training in deep neural networks, which achieved good performance in experiments. For future work, more studies inspired by biological rules will be employed to improve the perfromance of neural networks which are hard to train.

\section{Ackowledgments}
This work was jointly sponsored by the National Key Research and Development of China (No.2016YFB0800404) and the National Natural Science Foundation of China (No.61572068, No.61532005) and the Fundamental Research Funds for the Central Universities of China (No.2018YJS032).

\bibliographystyle{named}
\bibliography{arxiv}

\begin{thebibliography}{}

\bibitem[\protect\citeauthoryear{Bahdanau \bgroup \em et al.\egroup
  }{2014}]{Bahdanau2014Neural}
Dzmitry Bahdanau, Kyunghyun Cho, and Yoshua Bengio.
\newblock Neural machine translation by jointly learning to align and
  translate.
\newblock {\em Computer Science}, 2014.

\bibitem[\protect\citeauthoryear{Box and Pierce}{1968}]{Box1968Distribution}
G.~E.~P. Box and Davida. Pierce.
\newblock Distribution of residual autocorrelations in
  autoregressive-integrated moving average time series models.
\newblock {\em Publications of the American Statistical Association},
  65(332):1509--1526, 1968.

\bibitem[\protect\citeauthoryear{Cao \bgroup \em et al.\egroup
  }{2015}]{DBLP:conf/aaai/CaoHC15}
Wei Cao, Liang Hu, and Longbing Cao.
\newblock Deep modeling complex couplings within financial markets.
\newblock In {\em {AAAI} 2015}, pages 2518--2524, 2015.

\bibitem[\protect\citeauthoryear{Cho \bgroup \em et al.\egroup
  }{2014a}]{DBLP:conf/ssst/ChoMBB14}
Kyunghyun Cho, Bart van Merrienboer, Dzmitry Bahdanau, and Yoshua Bengio.
\newblock On the properties of neural machine translation: Encoder-decoder
  approaches.
\newblock In {\em EMNLP}, pages 103--111, 2014.

\bibitem[\protect\citeauthoryear{Cho \bgroup \em et al.\egroup
  }{2014b}]{DBLP:conf/emnlp/ChoMGBBSB14}
Kyunghyun Cho, Bart van Merrienboer, {\c{C}}aglar G{\"{u}}l{\c{c}}ehre, Dzmitry
  Bahdanau, Fethi Bougares, Holger Schwenk, and Yoshua Bengio.
\newblock Learning phrase representations using {RNN} encoder-decoder for
  statistical machine translation.
\newblock In {\em EMNLP}, pages 1724--1734, 2014.

\bibitem[\protect\citeauthoryear{{Conti} \bgroup \em et al.\egroup
  }{2017}]{2017arXiv171206560C}
E.~{Conti}, V.~{Madhavan}, F.~{Petroski Such}, J.~{Lehman}, K.~O. {Stanley},
  and J.~{Clune}.
\newblock {Improving Exploration in Evolution Strategies for Deep Reinforcement
  Learning via a Population of Novelty-Seeking Agents}.
\newblock arXiv:1712.06560, 2017.

\bibitem[\protect\citeauthoryear{Davoian and
  Lippe}{2007}]{DBLP:conf/dmin/DavoianL07}
Kristina Davoian and Wolfram{-}Manfred Lippe.
\newblock Time series prediction with parallel evolutionary artificial neural
  networks.
\newblock In {\em {ICDM} 2007}, pages 10--15, 2007.

\bibitem[\protect\citeauthoryear{Drucker \bgroup \em et al.\egroup
  }{1996}]{DBLP:conf/nips/DruckerBKSV96}
Harris Drucker, Christopher J.~C. Burges, Linda Kaufman, Alexander~J. Smola,
  and Vladimir Vapnik.
\newblock Support vector regression machines.
\newblock In {\em NIPS}, pages 155--161, 1996.

\bibitem[\protect\citeauthoryear{Du \bgroup \em et al.\egroup
  }{2015}]{DBLP:conf/cvpr/DuWW15}
Yong Du, Wei Wang, and Liang Wang.
\newblock Hierarchical recurrent neural network for skeleton based action
  recognition.
\newblock In {\em {CVPR} 2015}, pages 1110--1118, 2015.

\bibitem[\protect\citeauthoryear{Gowayyed \bgroup \em et al.\egroup
  }{2013}]{DBLP:conf/ijcai/GowayyedTHE13}
Mohammad~Abdelaziz Gowayyed, Marwan Torki, Mohamed~Elsayed Hussein, and Motaz
  El{-}Saban.
\newblock Histogram of oriented displacements {(HOD):} describing trajectories
  of human joints for action recognition.
\newblock In {\em {IJCAI} 2013}, pages 1351--1357, 2013.

\bibitem[\protect\citeauthoryear{Graves}{2012}]{DBLP:series/sci/2012-385}
Alex Graves.
\newblock {\em Supervised Sequence Labelling with Recurrent Neural Networks},
  volume 385 of {\em Studies in Computational Intelligence}.
\newblock Springer, 2012.

\bibitem[\protect\citeauthoryear{Hochreiter and
  Schmidhuber}{1997}]{DBLP:journals/neco/HochreiterS97}
Sepp Hochreiter and J{\"{u}}rgen Schmidhuber.
\newblock Long short-term memory.
\newblock {\em Neural Computation}, 9(8):1735--1780, 1997.

\bibitem[\protect\citeauthoryear{Holland}{1973}]{DBLP:journals/siamcomp/Holland73}
John~H. Holland.
\newblock Genetic algorithms and the optimal allocation of trials.
\newblock {\em {SIAM} J. Comput.}, 2(2):88--105, 1973.

\bibitem[\protect\citeauthoryear{Hulot \bgroup \em et al.\egroup
  }{2018}]{DBLP:conf/kdd/HulotAJ18}
Pierre Hulot, Daniel Aloise, and Sanjay~Dominik Jena.
\newblock Towards station-level demand prediction for effective rebalancing in
  bike-sharing systems.
\newblock In {\em Proceedings of the 24th {ACM} {SIGKDD} International
  Conference on Knowledge Discovery {\&} Data Mining, 2018}, pages 378--386,
  2018.

\bibitem[\protect\citeauthoryear{Ke \bgroup \em et al.\egroup
  }{2017}]{DBLP:conf/nips/KeMFWCMYL17}
Guolin Ke, Qi~Meng, Thomas Finley, Taifeng Wang, Wei Chen, Weidong Ma, Qiwei
  Ye, and Tie{-}Yan Liu.
\newblock Lightgbm: {A} highly efficient gradient boosting decision tree.
\newblock In {\em {NIPS} 2017}, pages 3149--3157, 2017.

\bibitem[\protect\citeauthoryear{Kim \bgroup \em et al.\egroup
  }{2017}]{DBLP:conf/icassp/KimHW17}
Suyoun Kim, Takaaki Hori, and Shinji Watanabe.
\newblock Joint ctc-attention based end-to-end speech recognition using
  multi-task learning.
\newblock In {\em ICASSP}, pages 4835--4839, 2017.

\bibitem[\protect\citeauthoryear{Kingma and
  Ba}{2014}]{DBLP:journals/corr/KingmaB14}
Diederik~P. Kingma and Jimmy Ba.
\newblock Adam: {A} method for stochastic optimization.
\newblock arXiv:1412.6980, 2014.

\bibitem[\protect\citeauthoryear{Lehman \bgroup \em et al.\egroup
  }{2017}]{Lehman2017Safe}
Joel Lehman, Jay Chen, Jeff Clune, and Kenneth~O. Stanley.
\newblock Safe mutations for deep and recurrent neural networks through output
  gradients.
\newblock arXiv:1712.06563, 2017.

\bibitem[\protect\citeauthoryear{Li and Bai}{2016}]{DBLP:conf/icmla/LiB16}
Xia Li and Ruibin Bai.
\newblock Freight vehicle travel time prediction using gradient boosting
  regression tree.
\newblock In {\em ICMLA}, pages 1010--1015, 2016.

\bibitem[\protect\citeauthoryear{Liang \bgroup \em et al.\egroup
  }{2015}]{Liang2015Assessing}
Xuan Liang, Tao Zou, Bin Guo, Shuo Li, Haozhe Zhang, Shuyi Zhang, Hui Huang,
  and Song~Xi Chen.
\newblock Assessing beijing's pm2. 5 pollution: severity, weather impact, apec
  and winter heating.
\newblock In {\em Proc. R. Soc. A}, volume 471, page 20150257, 2015.

\bibitem[\protect\citeauthoryear{Liang \bgroup \em et al.\egroup
  }{2018}]{DBLP:conf/ijcai/LiangKZYZ18}
Yuxuan Liang, Songyu Ke, Junbo Zhang, Xiuwen Yi, and Yu~Zheng.
\newblock Geoman: Multi-level attention networks for geo-sensory time series
  prediction.
\newblock In {\em {IJCAI} 2018}, pages 3428--3434, 2018.

\bibitem[\protect\citeauthoryear{Liu \bgroup \em et al.\egroup
  }{2018}]{DBLP:conf/aaai/LiuSZWT18}
Luchen Liu, Jianhao Shen, Ming Zhang, Zichang Wang, and Jian Tang.
\newblock Learning the joint representation of heterogeneous temporal events
  for clinical endpoint prediction.
\newblock In {\em {AAAI}, 2018}, 2018.

\bibitem[\protect\citeauthoryear{Lu \bgroup \em et al.\egroup
  }{2017}]{DBLP:conf/cvpr/LuXPS17}
Jiasen Lu, Caiming Xiong, Devi Parikh, and Richard Socher.
\newblock Knowing when to look: Adaptive attention via a visual sentinel for
  image captioning.
\newblock In {\em CVPR}, pages 3242--3250, 2017.

\bibitem[\protect\citeauthoryear{Plutowski \bgroup \em et al.\egroup
  }{1996}]{DBLP:journals/nn/PlutowskiCW96}
Mark Plutowski, Garrison~W. Cottrell, and Halbert White.
\newblock Experience with selecting exemplars from clean data.
\newblock {\em Neural Networks}, 9(2):273--294, 1996.

\bibitem[\protect\citeauthoryear{Qin \bgroup \em et al.\egroup
  }{2017}]{DBLP:conf/ijcai/QinSCCJC17}
Yao Qin, Dongjin Song, Haifeng Chen, Wei Cheng, Guofei Jiang, and Garrison~W.
  Cottrell.
\newblock A dual-stage attention-based recurrent neural network for time series
  prediction.
\newblock In {\em IJCAI}, pages 2627--2633, 2017.

\bibitem[\protect\citeauthoryear{Rumelhart \bgroup \em et al.\egroup
  }{1986}]{Rumelhart1986Learning}
David~E. Rumelhart, Geoffrey~E. Hinton, and Ronald~J. Williams.
\newblock Learning representations by back-propagating errors.
\newblock {\em Nature}, 323(6088):533--536, 1986.

\bibitem[\protect\citeauthoryear{Vemulapalli \bgroup \em et al.\egroup
  }{2014}]{DBLP:conf/cvpr/VemulapalliAC14}
Raviteja Vemulapalli, Felipe Arrate, and Rama Chellappa.
\newblock Human action recognition by representing 3d skeletons as points in a
  lie group.
\newblock In {\em {CVPR} 2014}, pages 588--595, 2014.

\bibitem[\protect\citeauthoryear{Yu \bgroup \em et al.\egroup
  }{2017}]{DBLP:conf/iccv/YuY0T17}
Zhou Yu, Jun Yu, Jianping Fan, and Dacheng Tao.
\newblock Multi-modal factorized bilinear pooling with co-attention learning
  for visual question answering.
\newblock In {\em ICCV}, pages 1839--1848, 2017.

\bibitem[\protect\citeauthoryear{Yule}{1927}]{Yule1927On}
G.~Udny Yule.
\newblock On a method of investigating periodicities in disturbed series, with
  special reference to wolfer's sunspot numbers.
\newblock {\em Philosophical Transactions of the Royal Society of London},
  226(226):267--298, 1927.

\bibitem[\protect\citeauthoryear{Zamora-Mart{\'\i}nez \bgroup \em et al.\egroup
  }{2014}]{zamora2014line}
F~Zamora-Mart{\'\i}nez, P~Romeu, P~Botella-Rocamora, and J~Pardo.
\newblock On-line learning of indoor temperature forecasting models towards
  energy efficiency.
\newblock {\em Energy and Buildings}, 83:162--172, 2014.

\bibitem[\protect\citeauthoryear{{Zhang} \bgroup \em et al.\egroup
  }{2017}]{2017arXiv171206564Z}
X.~{Zhang}, J.~{Clune}, and K.~O. {Stanley}.
\newblock {On the Relationship Between the OpenAI Evolution Strategy and
  Stochastic Gradient Descent}.
\newblock arXiv:1712.06564, 2017.

\end{thebibliography}

\end{document}